\documentclass[11pt]{article}

\usepackage[final]{acl}

\usepackage{times}
\usepackage{latexsym}
\usepackage{amsmath,booktabs,makecell}

\usepackage[T1]{fontenc}

\usepackage[utf8]{inputenc}

\usepackage{microtype}
\usepackage{amsmath}

\usepackage{inconsolata}

\usepackage{graphicx}

\usepackage{graphicx}
\usepackage{comment}
\usepackage{booktabs}
\usepackage{xcolor}
\usepackage{tcolorbox}
\usepackage{tikz}
\usepackage{multirow}
\usepackage{xspace}
\usepackage{url}

\newcommand{\benchmarkname}{\textit{ArcticAbstain}\xspace}
\newcommand{\datasetname}{\textit{ArcticQA}\xspace}

\usepackage{enumitem}

\title{Arctic Questions, Missing Answers: A Dataset and Benchmark for LLM Abstention in Arctic Science}

\author{
  \textbf{Benjamin Wilcox\textsuperscript{1}},
  \textbf{Dawei Gao\textsuperscript{1}},
  \textbf{Pradeeban Kathiravelu\textsuperscript{2}},\\
  \textbf{Douglas Causey\textsuperscript{2}},
  \textbf{Kewei Sha\textsuperscript{1}},\textbf{Yunhe Feng\textsuperscript{1}} \\
  \textsuperscript{1}University of North Texas, Denton, TX \\
  \textsuperscript{2}University of Alaska Anchorage, Anchorage, AK\\
  \texttt{benjamin.wilcox@unt.edu}, \texttt{dawei.gao@unt.edu}, \texttt{pkathiravelu@alaska.edu}\\ \texttt{dcausey@alaska.edu}, \texttt{kewei.sha@unt.edu}, \texttt{yunhe.feng@unt.edu}
}

\begin{document}
\maketitle

\begin{abstract}
Large language models (LLMs) should abstain from scientific multiple-choice questions when no option is valid, but frequent abstention alone does not demonstrate sensitivity to answer availability. We introduce \datasetname, a dataset of 194 questions derived from primary Arctic research, with automated checks of answer support and distractor contradiction against source evidence. We further develop \benchmarkname, a paired benchmark comparing answer-present and answer-absent conditions, with the correct answer replaced by a distractor in the latter and an explicit abstention option in both. We evaluate eight models from the Gemini, Claude, and ChatGPT families at high reasoning effort, with three trials per condition, yielding 9,312 recorded responses. Answer-present abstention rates range from 0.0\% to 63.0\%, whereas replacing the correct answer increases abstention by 5.05 percentage points on average. These findings highlight substantial baseline differences and the need to evaluate abstention frequency and responsiveness jointly. The dataset and benchmark are available at \url{https://github.com/BenWilcox8/arctic-qa}.
\end{abstract}

\section{Introduction}

Large language models (LLMs) could help researchers navigate and synthesize the multidisciplinary literature of Arctic science. Their usefulness, however, depends on whether they can provide scientifically supported answers and recognize when they should abstain. Arctic research spans interconnected physical, ecological, and human systems, and its findings often depend on specific locations, seasons, populations, and measurement conditions. Moreover, field observations are unevenly distributed across the region: prior research has documented substantial spatial biases in terrestrial Arctic sampling and scientific citation \citep{metcalfe2018patchy}. In this setting, an answer that is scientifically plausible in general may be incorrect within the scope of a particular question. Such errors can distort the interpretation of published findings and propagate into subsequent scientific analyses, making abstention an important component of trustworthy Arctic question answering (QA).

LLM abstention has already been investigated in other scientific disciplines. For example, \citet{wen2024characterizing} study abstention in biomedical and computer-science QA and show that behavior varies with the model, question type, and available context. Broader evaluations, including Abstain-QA and AbstentionBench, also reveal persistent difficulties in distinguishing questions that warrant an answer from those that warrant abstention \citep{madhusudhan2025dontanswer,kirichenko2025abstentionbench}. However, to the best of our knowledge, no dedicated benchmark has systematically evaluated LLM abstention on scientific questions grounded in primary Arctic research. Existing results therefore leave unresolved whether LLMs can appropriately withhold answers when the available choices conflict with the specific conditions and findings of Arctic studies.

Addressing this gap requires an evaluation resource that preserves the scope of Arctic scientific evidence. Geographic variation, seasonal dependence, and differences in study populations or measurement protocols complicate the construction of both correct answers and demonstrably incorrect alternatives. A finding from one setting cannot automatically establish what is valid in another. These considerations make benchmark curation particularly demanding: questions must state the relevant conditions, answers must be traceable to source findings, and distractors must be evaluated within the same scope. A dedicated Arctic resource is therefore necessary to examine abstention against explicit scientific evidence and to make the basis of the evaluation inspectable.

To support trustworthy evaluation, we introduce \datasetname{}, an evidence-grounded QA dataset constructed from primary Arctic research. For each item, we identify a clearly defined finding and its supporting evidence, then formulate a standalone question that preserves the conditions needed to interpret that finding. Separate automated checks assess whether the designated answer is supported by the source and whether each distractor is contradicted within the question's stated scope. Checking distractors explicitly is essential because an option's absence from a paper does not establish that it is false. Building on prior work in paper-grounded and automatically generated scientific QA \citep{dasigi-etal-2021-dataset,wan-etal-2024-sciqag}, \datasetname{} provides item-level evidence and validation checks to support transparent assessment of Arctic scientific answers.

Using \datasetname{}, we develop \benchmarkname{}, a paired benchmark centered on the following question: \emph{How does an LLM's abstention behavior change when a scientifically valid answer is removed from the available options?} We operationalize abstention as selecting an explicit abstention option instead of a substantive answer. In the \textit{answer-present} condition, each question includes one designated correct answer, three distractors, and the abstention option. In the matched \textit{answer-absent} condition, the correct answer is replaced with a fourth distractor, while the question and the remaining options are retained. The underlying scientific question remains answerable from its source; the manipulation changes whether a valid answer appears among the choices. Under this protocol, abstaining in the answer-present condition constitutes \emph{false abstention}, whereas selecting a substantive option in the answer-absent condition constitutes \emph{false commitment}. Comparing the two conditions measures sensitivity to answer availability relative to a model's baseline tendency to abstain, which a raw abstention rate alone cannot establish.

We evaluate eight models from the Gemini, Claude, and ChatGPT families on 194 Arctic scientific questions at high reasoning effort. Each model answers both conditions three times, yielding 9,312 recorded responses. We analyze abstention alongside the accuracy of non-abstaining responses. Answer-present abstention rates range from 0.0\% to 63.0\% across models, while replacing the correct answer increases abstention by 5.05 percentage points on average. These results reveal substantial variation in baseline abstention and a comparatively limited average change after answer replacement under our evaluation protocol. They underscore the need to evaluate both a model's willingness to answer Arctic scientific questions and its responsiveness to the absence of a valid option.

Our main contributions are as follows:
\begin{itemize}[nosep, leftmargin=*]
    \item \textbf{\datasetname{}: an evidence-grounded Arctic QA dataset.} We construct questions from primary Arctic research, retain item-level supporting evidence, and apply separate automated checks to answers and distractors to support trustworthy, inspectable evaluation.
    \item \textbf{\benchmarkname{}: a paired benchmark for Arctic scientific abstention.} We vary answer availability within matched question pairs to measure false abstention, false commitment, and changes in abstention relative to baseline behavior.
    \item \textbf{An empirical evaluation of eight LLMs.} We conduct repeated trials across three model families, characterizing differences in abstention behavior and sensitivity to answer replacement on Arctic scientific questions.
\end{itemize}

\section{Related Work}

Work on selective generation asks whether a model can estimate uncertainty and withhold an unreliable answer.
Self-evaluation prompts can improve selective generation \cite{ren2023selfevaluation}, while multi-model collaboration can expose knowledge gaps before a response is produced \cite{feng2024donthallucinate}.
Abstain-QA evaluates willingness to answer across answerable and unanswerable prompts \cite{madhusudhan2025dontanswer}.
AbstentionBench broadens this analysis across types of unanswerable questions and reports that reasoning-oriented training can increase forced problem solving \cite{kirichenko2025abstentionbench}.
These studies motivate our high-reasoning-effort setting, but their conclusions cannot be assumed to transfer directly to source-bounded Arctic scientific questions.

Scientific abstention has also been studied through controlled context changes and external verification.
Context-perturbation experiments alter the evidence available to a model and observe how its answer or abstention changes \cite{wen2024characterizing}.
Abstention-aware scientific reasoning systems can decompose claims and check them against retrieved evidence or natural-language-inference models \cite{abdaljalil2026knowing}.
Scientific claim-verification datasets such as SciFact-Open likewise connect claims to evidence retrieval and verification \cite{wadden-etal-2022-scifact}.
These systems evaluate the reliability of a composite process.
\datasetname instead evaluates the response of the tested model to a self-contained multiple-choice item, without giving the model a retrieval or verification pipeline.

Scientific question-answering resources cover difficult reasoning and questions anchored in research papers.
GPQA targets graduate-level questions in biology, physics, and chemistry, while MMLU-Pro expands difficult multi-task multiple-choice evaluation \cite{rein2023gpqa,wang2024mmlu}.
Other work builds information-seeking questions around research papers or generates scientific QA datasets with fine-grained checks \cite{dasigi-etal-2021-dataset,wan-etal-2024-sciqag}.
Round-trip consistency can filter synthetic question-answer pairs by testing whether an answer can be reconstructed \cite{alberti-etal-2019-synthetic}.
Decontextualization work formalizes the requirement that text remain understandable outside its original document \cite{choi-etal-2021-decontextualization}.
\datasetname combines these concerns by deriving each question from bounded source evidence.
The final question must also stand alone for a reader who does not have the paper.

Automatic multiple-choice construction introduces a separate risk: a plausible distractor can be partly correct, true under another scope, or equivalent to the answer.
Prior work therefore treats distractor generation and distractor evaluation as distinct tasks \cite{yu-etal-2024-automating-true,byun-choi-2025-gen,feng-etal-2024-exploring}.
Iterative critique can improve generated questions \cite{yao-etal-2025-mcqg}, but automated refinement and model judging can preserve shared errors or introduce judge bias \cite{zheng-etal-2023-judging,benedetto-etal-2025-survey}.
\datasetname responds with per-option source checks, deterministic rejection rules, bounded revision paths, and an explicitly non-gold machine-acceptance label.

\section{\datasetname Dataset}

We introduce \datasetname, a question-answering dataset grounded in specific findings from primary Arctic research. Reliable abstention evaluation requires both a supported correct answer and demonstrably incorrect distractors. We therefore develop a construction pipeline that links each target answer to supporting source evidence and verifies each distractor within the question's stated scope. The pipeline comprises paper selection, finding extraction and question generation, QA validation, and distractor generation and validation.

\subsection{Paper selection and eligibility}

We identified 84,829 papers through keyword searches using common Arctic research terms in Semantic Scholar\footnote{\url{https://www.semanticscholar.org/}}. Metadata filtering reduced this set to 4,420 candidate papers, whose full texts were subsequently assessed for eligibility. We defined geographic eligibility using a latitude threshold of $66.56^{\circ}$~N for terrestrial studies and a predefined list of eligible Arctic marine regions for marine studies. This list remained fixed throughout the eligibility assessment process.

\subsection{Finding extraction and question generation}

For each eligible paper, an LLM reads the full text and identifies the primary finding together with its supporting textual evidence. Once validated, the finding is held fixed as the target answer. A separate LLM writer then formulates a standalone question that asks specifically for this answer while preserving the scope and conditions of the original finding. This answer-first approach, followed by answer reconstruction, draws on the roundtrip-consistency principle of \citet{alberti-etal-2019-synthetic}. We adapt scientific-paper QA generation \citep{wan-etal-2024-sciqag} by restricting generation to one clearly scoped finding per paper.

\subsection{QA validation}

We validate each QA pair through blinded answer reconstruction followed by a separate verification step. The reconstructor LLM receives the question, its context, and the source evidence, but not the proposed answer. It produces a reconstructed answer, supporting evidence, and alternative answers. A separate verifier compares the proposed and reconstructed answers and assesses their consistency with the source. Specifically, it checks whether the evidence entails the proposed answer, whether the question and answer preserve the finding's scope and claim type, and whether the question's context is sufficient without disclosing the answer. A QA pair proceeds to distractor generation only after passing all checks. This combination of reconstruction and entailment checking is informed by the complementary factual-consistency signals identified by \citet{fabbri-etal-2022-qafacteval}.

\subsection{Distractor generation and validation}

For each validated QA pair, we construct four distractors through separate candidate-generation and verification stages. This staged design draws on prior work on automated distractor generation and filtering \citep{yu-etal-2024-automating-true,byun-choi-2025-gen}. A writer LLM proposes four to six candidates by modifying numerical values, categories, directions of change, scope, or entities. The writer does not verify its own outputs; each candidate undergoes a separate verification step.

To be retained, a candidate must be contradicted by specific source evidence within the question's stated scope. It is rejected if a reasonable interpretation of the paper could support it as a correct answer. Absence from the source alone does not establish falsity, and validity is assessed under the conditions specified by the question, including the relevant location and time. We retain four candidates that satisfy these criteria. These checks emphasize distractor validity, which prior work distinguishes from plausibility \citep{feng-etal-2024-exploring}, and make the basis for rejecting each option traceable to the source evidence.

\section{\benchmarkname Benchmark}

\benchmarkname evaluates LLM abstention on Arctic scientific questions using two matched multiple-choice conditions: \textit{answer-present} and \textit{answer-absent}. In the former, the correct answer is included among the options; in the latter, it is replaced by a validated distractor. Comparing responses across these conditions measures how abstention changes when a correct answer is no longer available for selection.

\subsection{Paired abstention protocol}

Each pair shares the same question wording and three validated distractors. Both items also include the explicit option ``I abstain from answering,'' consistent with prior work that allows LLMs to reject all candidate answers during self-evaluation \citep{ren2023selfevaluation}. Each item contains five options, with their order randomized to reduce systematic effects of option position.

\subsubsection{Answer-present condition}

The answer-present item contains the gold answer, three distractors, and the abstention option. Under the benchmark's scoring protocol, selecting the gold answer is correct, selecting a distractor is an incorrect answer, and selecting the abstention option constitutes \textit{false abstention}, or over-abstention. This condition quantifies how often a model abstains despite the presence of a valid answer.

\subsubsection{Answer-absent condition}

We replace the gold answer with a fourth validated distractor, retaining the same question, three shared distractors, and abstention option. Because no candidate answer is valid within the question's stated scope, abstention is the only response scored as correct. Selecting any distractor constitutes \textit{false commitment}, indicating failure to abstain when no valid answer is available.

The underlying question remains answerable from the source paper in both conditions; only the available answer options change. Together, the paired conditions distinguish false abstention from false commitment and quantify the change in abstention relative to the model's answer-present baseline behavior.

\section{Experiments}

\subsection{Experimental setup}

We evaluate eight models from three families: two Gemini models, three Claude models, and three ChatGPT models. All models are configured to use high reasoning effort. Each question from \datasetname is evaluated under the paired answer-present and answer-absent conditions defined in \benchmarkname. Each model answers each condition three times through separate calls, yielding six responses per model and 48 responses across all eight models for each base question.

The analysis includes 194 questions with all 48 planned response records, for a total of 9,312 recorded responses. Invalid responses are excluded from metric calculations; thus, the number of recorded responses is distinguished from the number of valid responses used in the metric denominators for reported results.

\subsection{Evaluation metrics}

We evaluate answer selection and abstention in \benchmarkname using six metrics from prior abstention research \cite{feng2024donthallucinate,wen2025knowyourlimits,wen2024characterizing}. For each model, we partition its responses across all evaluated questions and trials into five mutually exclusive categories. In the \textit{answer-present} condition, $N_1$, $N_2$, and $N_3$ denote the numbers of correct answers, incorrect answers, and abstentions, respectively. In the \textit{answer-absent} condition, $N_4$ and $N_5$ denote the numbers of incorrect answers and abstentions, respectively. Because no substantive option is correct in the answer-absent condition, every non-abstaining response in that condition is counted in $N_4$. Under the benchmark's answer-availability criterion, $N_3$ represents false abstentions and $N_5$ represents appropriate abstentions. The total number of evaluated responses is $N=\sum_{i=1}^{5}N_i$.

\textit{Abstention accuracy} (ACC) measures overall success in providing correct answers or abstaining when necessary:
\begin{equation}
    \mathrm{ACC}=\frac{N_1+N_5}{N}.
\end{equation}
It credits correct answers when a valid option is present and abstentions when all substantive options are incorrect, thereby jointly evaluating answer correctness and abstention behavior.

\textit{Abstention precision} measures the proportion of abstentions that occur when no valid answer option is available:
\begin{equation}
    \mathrm{Precision}_{\mathrm{abs}}=\frac{N_5}{N_3+N_5}.
\end{equation}
A higher value indicates that a greater proportion of the model's abstentions are appropriate under the evaluation protocol. Following the definition summarized by \citet{wen2025knowyourlimits}, \textit{abstention recall} is computed as
\begin{equation}
    \mathrm{Recall}_{\mathrm{abs}}=\frac{N_5}{N_2+N_4+N_5}.
\end{equation}
This metric measures the proportion of appropriate abstentions among the combined set of appropriate abstentions and incorrect substantive answers. Its denominator includes errors from both conditions: $N_2$ captures incorrect selections despite the availability of a valid option, and $N_4$ captures incorrect selections when no valid option is available. Thus, this definition penalizes incorrect answering in both conditions and differs from recall restricted to identifying answer-absent cases. We summarize abstention precision and recall using their harmonic mean, the \textit{abstention F1-score}:
\begin{equation}
    \mathrm{F1}_{\mathrm{abs}}=
    \frac{2\,\mathrm{Precision}_{\mathrm{abs}}\,
    \mathrm{Recall}_{\mathrm{abs}}}
    {\mathrm{Precision}_{\mathrm{abs}}+
    \mathrm{Recall}_{\mathrm{abs}}}.
\end{equation}

The \textit{abstention rate} measures how frequently the model abstains across both conditions:
\begin{equation}
    \mathrm{Abstention\ Rate}=\frac{N_3+N_5}{N}.
\end{equation}
This rate describes the model's overall tendency to withhold answers. Because it includes both false and appropriate abstentions, a higher abstention rate does not necessarily indicate better overall abstention performance.

Finally, \textit{reliable accuracy} (R-Acc) measures answer accuracy conditional on the model selecting a substantive option:
\begin{equation}
    \text{R-Acc}=\frac{N_1}{N_1+N_2+N_4}.
\end{equation}
Its denominator includes all non-abstaining responses across both conditions. R-Acc therefore characterizes the correctness of committed answers and should be interpreted alongside the abstention rate to account for how frequently the model chooses to answer.

\subsection{Aggregation and statistical analysis}

For the aggregate metric comparison, we compute each metric separately for the three trials and report the median across trials in Table~\ref{tab:final-metrics}. For the paired condition analysis in Table~\ref{tab:condition-shifts}, we pool responses across all three trials to estimate each model's condition-specific abstention rates. The paired shift is the answer-absent rate minus the answer-present rate, expressed in percentage points (pp).

We estimate 95\% confidence intervals for these shifts using 10,000 bootstrap resamples of questions with a random seed of 7. Each resample retains both conditions and all three trials for every selected question, preserving the dependence among responses associated with that question. For each model, a question-level sign-flip permutation test assesses whether abstention changes after the correct answer is replaced. We apply Holm correction across the eight model-specific tests to control the family-wise error rate and assess significance at an adjusted $p<0.05$.

\subsection{Results and analysis}

\subsubsection{Variation in baseline abstention}

Table~\ref{tab:condition-shifts} shows substantial variation in answer-present abstention, ranging from 0.0\% to 63.0\%. ChatGPT 5.6 Sol and ChatGPT 5.6 Terra almost always select a substantive option, with abstention rates of 0.2\% and 0.0\%, respectively. The Gemini models abstain on 6.0\%--6.4\% of answer-present responses, whereas the Claude models abstain on 36.1\%--48.5\%. ChatGPT Astra has the highest rate, at 63.0\%. The large differences among the three ChatGPT models show that model-family membership alone does not characterize abstention behavior in this evaluation.

\begin{table*}[!t]
    \centering
    \fontsize{10}{12}\selectfont
    \setlength{\tabcolsep}{4pt}
    \renewcommand{\arraystretch}{1.2}
    \setlength{\abovecaptionskip}{6pt}
    \begin{tabular*}{\textwidth}{@{\extracolsep{\fill}}lccccc@{}}
        \toprule
        \textbf{Model}
        & \textbf{Present abst.}
        & \textbf{Absent abst.}
        & \textbf{Shift [95\% CI] (pp)}
        & \textbf{Permutation} $\boldsymbol{p}$
        & \textbf{Holm} $\boldsymbol{p}$ \\
        \midrule
        Gemini 3.8 Flash
        & 6.4\% & 10.3\% & $+3.9\;[0.6,\,7.4]$
        & 0.037 & 0.187 \\
        Gemini 3.7 Flash
        & 6.0\% & 8.6\% & $+2.6\;[-0.1,\,5.4]$
        & 0.063 & 0.198 \\
        \addlinespace[3pt]
        Claude Fable 5.1
        & 36.1\% & 46.9\% & $+10.8\;[5.7,\,16.1]$
        & 0.00010 & 0.0008 \\
        Claude Opus 5
        & 41.3\% & 47.5\% & $+6.2\;[0.9,\,11.6]$
        & 0.031 & 0.187 \\
        Claude Sonnet 5
        & 48.5\% & 51.0\% & $+2.5\;[-1.7,\,6.9]$
        & 0.331 & 0.331 \\
        \addlinespace[3pt]
        ChatGPT Astra
        & 63.0\% & 74.1\% & $+11.1\;[5.7,\,16.6]$
        & 0.00015 & 0.0011 \\
        ChatGPT 5.6 Sol
        & 0.2\% & 2.1\% & $+1.9\;[0.3,\,4.0]$
        & 0.049 & 0.198 \\
        ChatGPT 5.6 Terra
        & 0.0\% & 1.5\% & $+1.5\;[0.3,\,3.3]$
        & 0.065 & 0.198 \\
        \bottomrule
    \end{tabular*}
    \caption{Condition-specific abstention rates pooled across three trials for 194 questions. Shifts are answer-absent minus answer-present rates, in percentage points (pp). The 95\% confidence intervals use question-level bootstrap resampling. Permutation tests use question-level paired differences; Holm-adjusted $p$-values account for the eight model-specific tests.}
    \label{tab:condition-shifts}
\end{table*}

\subsubsection{Response to correct-answer removal}

\begin{figure}[t]
    \centering
    \includegraphics[width=\columnwidth]{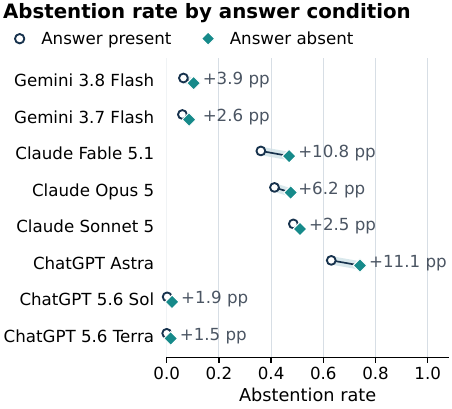}
    \caption{Abstention rates under the answer-present and answer-absent conditions for the eight models. Each segment connects a model's two rates, pooled across three trials. Labels report the answer-absent minus answer-present shift in percentage points.}
    \label{fig:high-abstention-shift}
\end{figure}

Replacing the correct answer with a distractor increases the abstention point estimate for every model. The increases range from 1.5 to 11.1 percentage points, with a mean increase of 5.05 percentage points across the eight models. These shifts are modest relative to the 63.0\% range in answer-present abstention. Figure~\ref{fig:high-abstention-shift} visualizes each model's paired rates.

Claude Fable 5.1 and ChatGPT Astra exhibit the largest increases, at 10.8 and 11.1 percentage points, respectively. Both changes remain statistically significant after Holm correction, with adjusted $p$-values of 0.0008 and 0.0011. The remaining six models do not meet the adjusted significance threshold. Thus, the point estimates consistently indicate increased abstention after answer removal, while the model-specific tests provide evidence of a change for two models after correction for multiple comparisons.

\subsubsection{Accuracy and abstention must be interpreted jointly}

Table~\ref{tab:final-metrics} reports median metrics across the three trials. ChatGPT Astra has the highest median ACC (0.541), abstention F1-score (0.618), and R-Acc (0.540), together with the highest overall abstention rate (0.686). However, its 63.0\% answer-present abstention rate also indicates frequent abstention when a valid option is available. Its comparatively high conditional answer accuracy therefore coexists with substantial false abstention.

All reported R-Acc medians exceed 0.125, the expected value for uniform random selection among the four substantive options under equally represented conditions: $\tfrac{1}{2}\times\tfrac{1}{4}=0.125$. The other seven models have R-Acc medians between 0.261 and 0.379. These descriptive comparisons do not establish statistically significant differences between models. For example, ChatGPT 5.6 Sol has an R-Acc of 0.291, numerically close to the Gemini models' values of 0.303 and 0.306, but a much lower abstention rate of 0.011. Conditional answer accuracy and the tendency to abstain therefore capture distinct aspects of performance, and both are needed to assess reliability on Arctic scientific questions.

\begin{table}[!t]
    \centering
    \fontsize{10}{12}\selectfont
    \setlength{\tabcolsep}{2pt}
    \renewcommand{\arraystretch}{1.18}
    \setlength{\abovecaptionskip}{6pt}
    \begin{tabular*}{\columnwidth}{@{\extracolsep{\fill}}lcccc@{}}
        \toprule
        \textbf{Model} & \textbf{ACC} & \shortstack{\textbf{Abst.}\\\textbf{rate}}
        & $\mathbf{F1}_{\mathbf{abs}}$ & \textbf{R-Acc} \\
        \midrule
        Gemini 3.8 Flash   & 0.329 & 0.084 & 0.134 & 0.303 \\
        Gemini 3.7 Flash   & 0.326 & 0.073 & 0.113 & 0.306 \\
        \addlinespace[2pt]
        Claude Fable 5.1   & 0.458 & 0.415 & 0.465 & 0.379 \\
        Claude Opus 5     & 0.438 & 0.444 & 0.459 & 0.359 \\
        Claude Sonnet 5   & 0.408 & 0.498 & 0.464 & 0.303 \\
        \addlinespace[2pt]
        ChatGPT Astra     & 0.541 & 0.686 & 0.618 & 0.540 \\
        ChatGPT 5.6 Sol   & 0.298 & 0.011 & 0.029 & 0.291 \\
        ChatGPT 5.6 Terra & 0.266 & 0.008 & 0.021 & 0.261 \\
        \bottomrule
    \end{tabular*}
    \caption{Metrics pooled across the three trials for eight models evaluated at high reasoning effort on 194 Arctic scientific questions. Each trial's metrics combine the answer-present and answer-absent conditions.}
    \label{tab:final-metrics}
\end{table}

\section{Conclusion}

We introduced \datasetname, a dataset grounded in primary Arctic research, and \benchmarkname, a paired benchmark for evaluating LLM abstention under answer-present and answer-absent conditions. Across eight models, answer-present abstention rates ranged from 0.0\% to 63.0\%, while removing the correct answer increased abstention by 5.05 percentage points on average. These findings highlight the distinction between a model's tendency to abstain and its responsiveness to answer availability. Reliable evaluation should therefore consider false abstention, false commitment, and the accuracy of non-abstaining responses jointly. Our dataset and benchmark provide a foundation for assessing and improving LLM reliability in Arctic scientific question answering.

\section{Limitations}

\benchmarkname is a geographically bounded case study whose coverage reflects discovery queries, full-text availability, and model-guided processing order. It is not representative of all Arctic research or other scientific domains. Dataset construction relies on model-generated content and automated validation. No sample was reviewed by domain experts; thus, \texttt{machine\_accepted\_unverified} denotes automated acceptance, not expert-verified ground truth. The residual error rate remains unmeasured. The evaluation covers 194 questions, eight models, a single prompt, one explicit abstention option, and high reasoning effort. The paired multiple-choice task assesses rejection of invalid option sets rather than spontaneous abstention in open-ended scientific work. The results do not establish training exposure, explain observed response policies, or demonstrate generalization beyond the evaluated setting.

\bibliography{reference}

\end{document}